\documentclass[conference]{IEEEtran}
\IEEEoverridecommandlockouts
\usepackage{cite}
\usepackage{amsmath,amssymb,amsfonts}
\usepackage[linesnumbered,ruled,vlined]{algorithm2e}
\usepackage{algpseudocode}
\usepackage{graphicx}
\usepackage{textcomp}
\usepackage{multirow}
\usepackage{xcolor}
\usepackage{bm}
\usepackage{cite}
\usepackage{amsmath,amssymb,amsfonts}

\usepackage{textcomp}
\usepackage{xcolor}
\def\BibTeX{{\rm B\kern-.05em{\sc i\kern-.025em b}\kern-.08em
    T\kern-.1667em\lower.7ex\hbox{E}\kern-.125emX}}
\begin{document}

\title{Assessing Personalized AI Mentoring with Large Language Models in the Computing Field\\
}

\author{
   \IEEEauthorblockN{Xiao Luo\IEEEauthorrefmark{1}, Sean O'Connell \IEEEauthorrefmark{2}, Mithun, Shamima \IEEEauthorrefmark{2}}
    \IEEEauthorblockA{\IEEEauthorrefmark{1} Department of Management Science and Information Systems, Oklahoma State University, USA
    \\xiao.luo@okstate.edu}
    \IEEEauthorblockA{\IEEEauthorrefmark{2} Department of Computer Information Technology, Purdue University, Indianapolis, USA
    \\\{oconne35, smithun\}@purdue.edu}
   
}

\maketitle

\begin{abstract}
This paper provides an in-depth evaluation of three state-of-the-art Large Language Models (LLMs) for personalized career mentoring in the computing field, using three distinct student profiles that consider gender, race, and professional levels. We evaluated the performance of GPT-4, LLaMA 3, and Palm 2 using a zero-shot learning approach without human intervention. A quantitative evaluation was conducted through a custom natural language processing analytics pipeline to highlight the uniqueness of the responses and to identify words reflecting each student's profile, including race, gender, or professional level. The analysis of frequently used words in the responses indicates that GPT-4 offers more personalized mentoring compared to the other two LLMs. Additionally, a qualitative evaluation was performed to see if human experts reached similar conclusions. The analysis of survey responses shows that GPT-4 outperformed the other two LLMs in delivering more accurate and useful mentoring while addressing specific challenges with encouragement languages. Our work establishes a foundation for developing personalized mentoring tools based on LLMs, incorporating human mentors in the process to deliver a more impactful and tailored mentoring experience.
\end{abstract}


\section{Introduction}

AI mentoring is increasingly recognized as a critical tool in education, offering personalized guidance and support to students in ways that traditional mentoring may not always provide. The importance of AI-driven mentoring lies in its ability to scale mentoring efforts, making high-quality support accessible to a larger number of students, particularly in fields like computing where individual mentorship demand often exceeds supply \cite{roll2016evolution}. AI mentors can offer personalized feedback, identify student strengths and weaknesses, and adapt to individual learning styles, fostering a more tailored educational experience \cite{holstein2018classroom}. Additionally, AI mentors are available around the clock, providing assistance whenever students need it, thus enhancing the learning process \cite{kumar2010architecture}. Recent trends show a growing integration of AI in mentoring systems, with advancements in natural language processing and machine learning enabling more nuanced and context-aware interactions \cite{gavsevic2015let,frankford2024ai}. As these technologies evolve, AI mentoring is expected to become more sophisticated, offering increasingly effective support for students' academic and personal development.

The trend of AI mentoring using Large Language Models (LLMs) is rapidly gaining momentum, driven by the ability of these models to understand and generate human-like text across a wide range of contexts. LLMs, such as OpenAI GPT, Meta LLaMA, and Google Gemini, etc., have shown remarkable capability in natural language understanding, making them ideal candidates for mentoring and educational applications \cite{benitez2024harnessing}. These models can engage in meaningful dialogue, provide personalized feedback, and support students in navigating complex subject matter, particularly in domains like computing where mentorship is crucial \cite{xu2024large}. The ability of LLMs to process and analyze vast amounts of textual data allows them to tailor responses based on individual student needs and background, fostering a more personalized learning experience through textual data \cite{fagbohun2024beyond}. Moreover, the integration of LLMs in educational tools is becoming more prevalent, as they are increasingly being used to develop intelligent tutoring systems that can simulate one-on-one mentoring sessions in various domains \cite{abd2023large}. As LLMs continue to evolve, their role in AI mentoring is expected to expand, offering even more sophisticated and contextually aware support to students across various disciplines.


In this research, we evaluated current state-of-the-art Large Language Models (LLMs) about their effectiveness in personalized mentoring for students in computing domain. We focused evaluating the ability of LLMs on providing personalized mentoring experiences for career planning in the computing fields while considering mentees' social backgrounds and proficiency levels in computing. The research questions guiding our investigation are following: 
\begin{itemize}
    \item RQ1: Considering the low enrollment of underrepresented students in computing programs \cite{lehman2021growing}, it's important to recognize that students from diverse social backgrounds may have varying needs. Our first research question is: If LLMs are utilized for personalized mentoring, do they take into account the student's social background (e.g., race, ethnicity)?
    \item RQ2: Considering that computing students have varying levels of experience and education in college, their needs may differ. Our second research question is: If LLMs are employed for personalized mentoring, do they take into account the student's educational background (e.g., freshman, sophomore)?
\end{itemize}

In our study, we gathered 15 questions from computing major students with diverse social and educational backgrounds. These questions focused on their interest in computing careers. We then evaluated three state-of-the-art LLMs to assess their ability to provide personalized mentoring by answering these questions. To assess the quality of the personalized responses to these mentoring questions, we employed both quantitative and qualitative evaluation methods. The quantitative evaluation utilized a developed natural language processing pipeline to analyze how well the LLMs' answers accounted for the students' social and educational backgrounds. In contrast, the qualitative evaluation focused on analyzing the usefulness of the responses.

To the best of our knowledge, this research is the first to evaluate the personalized AI mentoring using LLMs by considering both social and educational backgrounds of the students. The main contributions of this paper include:

\begin{itemize}
    \item Conduct a systematic evaluation of LLMs for personalized AI mentoring in the computing domain, specifically focusing on career planning.
    \item Highlight the strengths and limitations of LLMs in delivering personalized mentoring.
    \item Demonstrate potential future advancements in utilizing LLMs for personalized mentoring.
\end{itemize}

\section{Related Work}
\subsection{LLMs in Education}

GPT-based models have been actively explored in various educational fields, including academic writing \cite{buruk2023academic,zohery2023chatgpt}, teaching introductory programming \cite{biswas2023role,pankiewicz2023large}, and mathematics to impart fundamental skills \cite{wardat2023chatgpt, supriyadi2023exploring}. 
Recently, there has been a growing application of AI tools in computer science and information technology education \cite{yilmaz2023effect, wang2023exploring}. These AI-based tools are being utilized to teach introductory programming courses, offering the capability to automatically generate code, identify errors, and provide suggestions to help students produce precise and efficient code \cite{  kosar2024computer,  yilmaz2023augmented,hartley2024artificial, xue2024does, haindl2024students}. Despite the potential of LLM-based models, research on the effectiveness of tools like ChatGPT is limited. 
Okonkwo and Ade-Ibijola \cite{okonkwo2022revision} highlighted the transformative nature of AI tools in traditional teaching methods, offering personalized learning and automated tasks. AI-based chatbots were specifically mentioned for their role in problem-solving and personalized guidance. However, Kosar et al. \cite{kosar2024computer} pointed out challenges with AI chatbots, such as fostering laziness and hindering critical thinking. Ismail and Ade-Ibijola \cite{ismail2019lecturer} suggested designing AI chatbots that prioritize emotional and personalized engagement to help students with programming difficulties, emphasizing continuous practice over memorization to reinforce a deeper understanding of programming principles. In conclusion, the literature consistently indicates the significant potential of AI tools in computing education. 

\subsection{LLMs for Mentoring}

Several review articles \cite{rane2024enhancing, bagai2024designing} have examined the potential and challenges of AI-enhanced personalized mentoring. Bagai and Mane \cite{bagai2024designing} emphasize the promising possibilities and challenges, indicating that AI-enabled mentorship could improve career advancement, skill development, and mentee satisfaction. However, concerns such as security, algorithmic bias, and ethical considerations persist. The article also covers the essential characteristics and technological foundations for effective AI mentoring platforms, concluding that a fully realized AI-driven mentorship platform is still under development. Cronjé’s 2023 study \cite{cronje2023exploring} utilized ChatGPT to assist fourth-year IT students with research proposals. Key findings emphasized the quality of feedback, the importance of well-designed prompts, and the necessity of student reflection. The study highlighted that well-designed prompts and reflection are vital for effective AI interaction. Akiba and Fraboni’s 2023 study \cite{akiba2023ai} investigates the potential of AI-powered tools like ChatGPT to improve the accessibility, efficiency, and effectiveness of academic advising. Similar to our research, the authors compiled a list of frequently asked questions from current and prospective students in a teacher education bachelor’s degree program in the United States, selecting seven for this study. These questions were input into the ChatGPT to evaluate the quality and delivery of the generated answers. 
Sanya-Isijola and Leung \cite{sanya2024chatgpt} found ChatGPT to be valuable for learning, collaboration, exam preparation, and keeping updated. Other studies \cite{limo2023personalized, thomasusing} reported positive outcomes for AI-enhanced mentors and stressed the necessity of human involvement to enhance AI performance.


To the best of our knowledge, no existing work uses both NLP techniques and human evaluation to compare the performance of multiple LLMs for personalized mentoring towards career planning in computing field. Our study is the first to evaluate personalized mentoring while taking into account the mentees' social backgrounds and experience levels. 

\begin{figure*}[h!]
    \centering
    \includegraphics[width=0.8\textwidth]{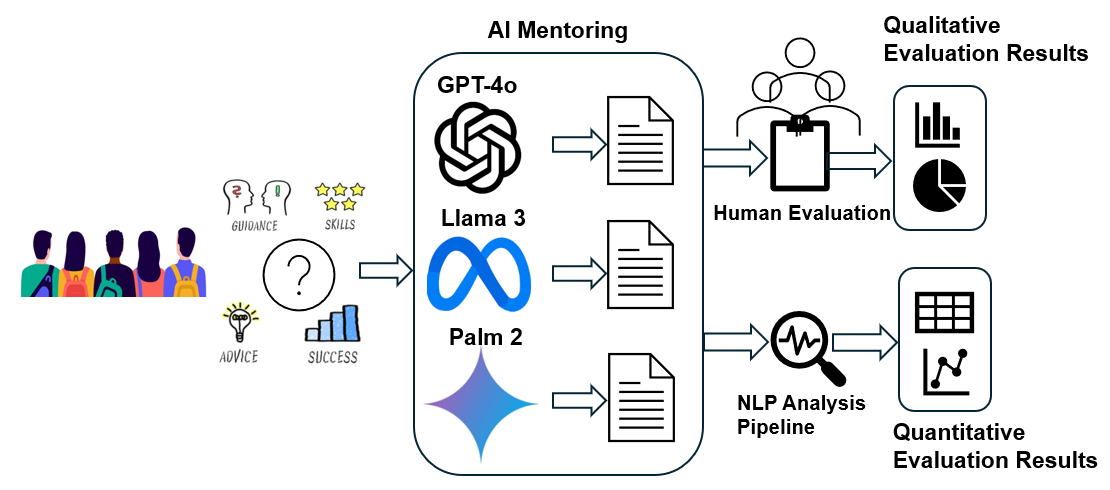}
    \caption{Overview of the AI mentoring and analysis using LLMs}
    \label{fig:system}
\end{figure*}

\section{Methodology}
In this research, we utilized a proposed personalized AI mentoring framework, as illustrated in Figure \ref{fig:system}. The framework takes various questions from students and generates responses. Three LLMs were incorporated into this framework for evaluation purposes. We then evaluated these LLMs using a developed NLP pipeline, beginning with an analysis of the similarities between the responses to each question, considering different student social and educational backgrounds. Following this, we focused on identifying the unique topics within the answers for each individual question. Finally, we conducted an independent qualitative analysis of the responses generated by each LLM for each question.

\subsection{Student Profile and Mentoring Questions}
To comprehensively assess the capability of LLMs for personalized AI mentoring, we created three distinct student profiles and prepared 15 mentoring questions. Table I shows the student profiles included in this research, which covers both genders, three different race/ethnicity backgrounds, two different levels of college experiences, and two different majors. 

\begin{table}[h!]
    \centering \caption{Student Profiles Designed for Evaluating Personalized AI Mentoring} \label{tab:my_label}
    \setlength{\tabcolsep}{1.5pt}
    \begin{tabular}{c|c|c|c|c}
    
    \hline
      Profile \# &  Gender & Race/Ethnicity & College Year & Major \\\hline
    M-AA-J-CS&    Male & African American & Junior & Computer Science\\\hline
    M-W-F-U &    Male & White & Freshman & Undecided\\\hline
     F-H-F-CS & Female & Hispanic & Freshman & Computer Science\\\hline
    \end{tabular}
\end{table}

Table \ref{tab:my_label} presents the list of questions included in this study. In our previous research \cite{mithun2024designing}, we conducted a survey to assess the mentoring needs of students. Participants included first-year and third-year computing students. We analyzed their feedback, organizing it into common themes such as resumes, internships, and networking. For the current study, we selected 15 questions to address the various themes identified in the students' responses. We also categorized these questions to determine if personalized responses are necessary. Personalized responses might vary based on the students' social (e.g., race, ethnicity) or educational background (e.g., freshman, sophomore, undecided major, computer science major). For instance, for a question like ``Please advise me on future computing career plans and how to proceed," the response should consider whether the student is a junior or freshman, or if they have decided to major in computer science. Responses aimed at junior computer science students would focus more on staying updated with current technologies and industrial networking, etc. 

\begin{table*}[h!]
    \centering \caption{Mentoring Questions based on Mentoring Needs} \label{tab:my_label}
    \setlength{\tabcolsep}{1.5pt}
    \begin{tabular}{p{1cm}|p{16.8cm}}
    \hline
    Question & Content \\\hline   
     \multicolumn{2}{l}{Personalized Response Expected} \\\hline  
      1.  & Please advise me on future computing career plans and how to proceed. \\\hline 
         2.  & What are the most common struggles/cons of doing computing work? \\\hline 
  6.	& What are employers looking for in a resume for computing jobs? \\\hline

 8.	& In the computing field, how can I network? How do I approach people? How will they remember me? \\\hline
9.	&How do people deal with stress in computing jobs?\\\hline
10.	&What is the work-life balance like in computing fields? \\\hline
 13. & How can I improve my knowledge in the computing career?  How can I stand out compared to other candidates? \\\hline
12. & What are the biggest challenges for finding the right job in computing? How can job-seekers overcome those challenges?\\\hline
15. & I would like to know more about what skills I should be developing right now to make myself more marketable for internships and employment in the computing field. What is most important right now to get my first internship as a beginner? \\\hline

\multicolumn{2}{l}{Generic Response Expected} \\\hline  

      3.	& What I want to know about a computing career is whether many jobs are available and how much the positions in this career pay.\\\hline
 4.	& How many years of education/training should I expect to get a computing job?\\\hline
 5. & 	I want to know what the work environment is like in computing careers. I am curious as to what the computing profession looks like in the field. I'm interested in what they work on and what a day-to-day looks like. \\\hline
7.	 &I would like to learn more about internships. How to get them, what to expect, etc. in computing fields. \\\hline
11. &  I am interested in learning about computing careers and required education plans.\\\hline

14. & How can I improve my coding skills? \\\hline
    \end{tabular}
\end{table*}

\subsection{AI Mentoring using LLMs}
In this research, we systematically evaluated these three LLMs towards personalized AI mentoring through conducting the zero-shot experiments using the Application Programming Interfaces (APIs) of GPT-4o \cite{achiam2023gpt}, PaLM 2 \cite{anil2023palm}, and LLaMA 3 \cite{dubey2024llama}. The API allows users to provide instructions via two role variables. Our prompt is structured in the following order:

\begin{itemize}
    \item  System: defines task instructions for LLMs in the desired role. We use
the system variable to provide task instructions so that the model acts as the role of
an AI mentor and provide suggestions to students who asked question with social and educational background information.
\item User: seek mentor suggestion by inputting question and social and educational background. 
\end{itemize}

Following the above definitions, a user message is designed to contain a description of student's background information and the question relevant to computing career planning. Figure 2 illustrates the design of the prompt. The prompt guides the LLMs to consider the student's background information when providing mentoring suggestions.

\begin{figure}[h!]
    \centering
    \includegraphics[width=0.45\textwidth]{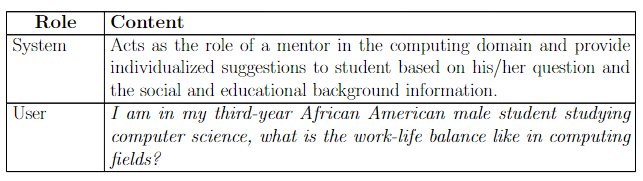}
    \caption{Design of the prompt for AI mentoring}
    \label{fig:prompt}
\end{figure}

\subsection{NLP Analysis Pipeline}

The NLP analysis pipeline designed in this research is to objectively evaluate whether and how the LLMs consider each student's social and educational background when answering the questions. 

The pipeline first analyzes the similarity between the answers for each question using each LLM. The hypothesis is that the percentage of sentences in the answers with high semantic similarity ($>=\theta$) need to be low to have distinct answers. For example, if the answers of the same question from two students with different background have more 90\% of the sentences with high semantic similarity, they are treated as the same answer, which means the AI mentoring does not provider personalized answers. To calculate the semantic similarity, we applied sentence transformer to convert candidates and ground truth into embeddings, then, applied cosine similarity to measure the similarity shown as Equation 1. Here, $a\cdot b$ represents the dot product of embeddings of $a$ and $b$, and $||a||$ and $||b||$ represent the magnitudes (or Euclidean norms) of embeddings of $a$ and $b$, respectively.

\begin{equation}\label{nlvt}
c(a, b) = \frac{\mathbf{a} \cdot \mathbf{b}}{\|\mathbf{a}\| \|\mathbf{b}\|}
\end{equation}

We computed the percentage of sentences in the responses that had a high semantic similarity, using a threshold of 0.7. This threshold indicates that if two sentences have a cosine similarity greater than 0.7, they are considered to have similar or identical semantic meanings and are counted as the same sentence.

After conducting a similarity analysis, we extracted topics from each response and identified both shared and unique topics among the answers. We then explored whether the unique aspects were influenced by technical skills, social skills, or cultural or community perspectives.

\subsection{Design of Human Evaluation}
\begin{figure*}[h!]
    \centering
    \includegraphics[width=0.8\textwidth]{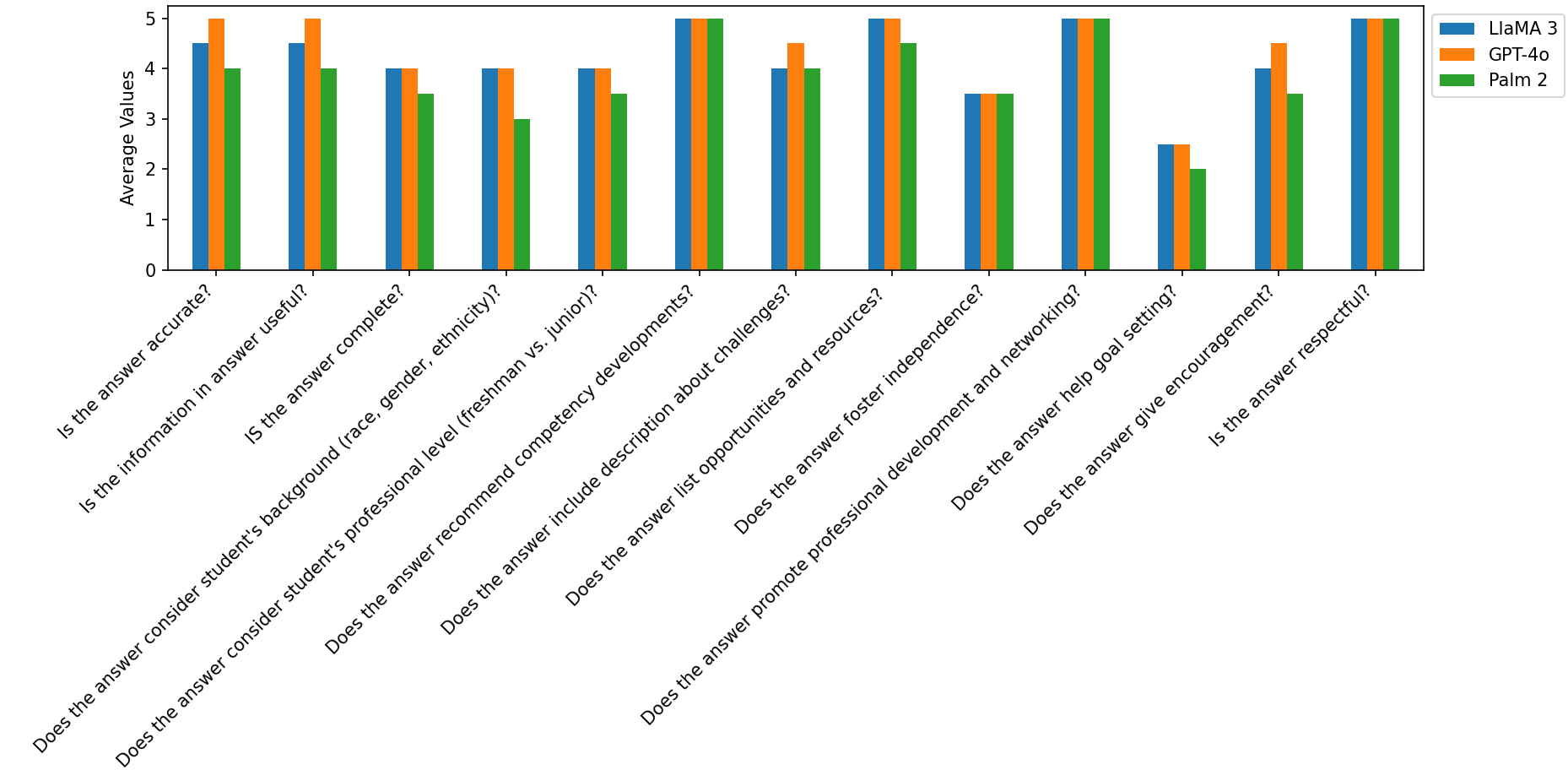}
    \caption{Human Evaluation of the LLMs towards Mentoring}
    \label{fig:mentoringhuman}
\end{figure*}

Other than the objective analysis using the NLP analysis pipeline, we developed a protocol for human evaluation. Specifically, we designed a survey with thirteen questions (5-point Likert scale with 1 being completely disagreed and 5 being completely agreed) based on the existing evaluation metrics used to assess the responses of a generative AI \cite{goodman2023accuracy,akiba2023ai} as well as other criteria used to evaluate human mentors' effectiveness \cite{yukawa2020new,berk2005measuring} that considering whether it is supporting mentees academically and emotionally to reach to their career goals. The designed survey also measured whether the responses considered the unique social and educational backgrounds of the students who asked the mentoring questions. Figure  \ref{fig:mentoringhuman} shows the evaluation questionnaires along with the results.
\begin{table*}[h!]\label{tab:zero_shot}
 \setlength{\tabcolsep}{1.5pt}
    \centering
    \caption{LLM Performance Comparison Based on Semantic Similarity Analysis}
    \begin{tabular}{|p{2cm}|c|c|c|c|c|c|c|c|c|c|}\hline
    
    \multirow{2}{*}{Question} & \multicolumn{3}{c|}{GPT-4o} & \multicolumn{3}{c|}{LLaMA 3} & \multicolumn{3}{c|}{Palm 2} \\\cline{2-10}
    & M-AA-J-CS & M-W-F-U & F-H-F-CS & M-AA-J-CS & M-W-F-U & F-H-F-CS & M-AA-J-CS & M-W-F-U & F-H-F-CS\\\hline
 \multicolumn{10}{|l|}{Personalized Responses (PR)} \\\hline  
    1 & 0.368 &	0.118 &	0.412 & 0.250 & 	0.161 &	0.244 & 0.471 &	0.250 &	0.333 \\\hline
    2 & 0.273 &	0.087 &	0.263 & 0.125 &	0 &	0.100 & 0.200 &	0.300 &	0.357 \\\hline
     6 & 0.294 &	0.368 &	0.381 & 0.375 & 0.261 & 0.750 & 0.500 &	0.087 &	0.145 \\\hline
   
    8 & 0.121 &	0.135 &	0.411 & 0.636 & 0.167 &	0.325 & 0.823 &	0.625 &	0.625 \\\hline
    9 & 0.347 &	0.421 &	0.473 & 0.500 &	0.136 &	0.348 & 0.895 &	0.211 &	0.417 \\\hline
     10 & 0 &	0.267 &	0.267 & 0.400 &	0.178 &	0.367 & 0.579 &	0.333 &	0.706 \\\hline
         12 & 0.4& 0.028 &	0.360 & 0.261 & 0.022 &	0.240 & 0.350 &	0.053 &	0.261 \\\hline
     13 & 0.286 &	0.160 &	0.473 & 0.333 &	0.163 &	0.444 & 0.556 &	0.071 &	0.313 \\\hline
   
    15 & 0.381 &	0.174 &	0.414 & 0.555& 	0.154 &	0.342 & 0.364 &	0.307 &	0.454 \\\hline
    PR-Average & 0.274	&0.195	&0.384	&0.382	&0.138&	0.351&	0.526	&0.249	&0.401

	\\\hline
    \multicolumn{10}{|l|}{Generic Responses (GR)} \\\hline  
    
    3 & 0.529 &	0.214 &	0.176 & 0.142 &	0.193 &	0.346 & 0.182 &	0.175 &	0.320  \\\hline
    4 & 0.385 &	0.400 &	0.714 & 0.043 &	0.024 &	0.118 & 0.667 &	0.222 &	0.412 \\\hline
    5 & 0.600 &	0.217 &	0.579 & 0.238 &	0.060 &	0.150 & 0.500 &	0.105 &	0.381 \\\hline
    7 & 0.480 &	0.190 &	0.522 & 0.368 & 0.250 &	0.400 & 0.645 &	0.452 &	0.686 \\\hline
   
    11 & 0.474 &	0.038 &	0.471 & 0.325 &	0.192 &	0.310 & 0.823 &	0.212 &	0.4 \\\hline

     14 & 0.348 &	0.130 &	0.391 & 0.348 &	0.167 &	0.444 & 0.571 &	0.294 &	0.411 \\\hline
    GR-Average & 0.469&	0.198	&0.476	&0.244&	0.148	&0.295&	0.565	&0.243	&0.435

\\\hline
    Overall Average & 0.352 &	0.197 &	0.421 & 0.327 &	0.142 &	0.328 & 0.542 &	0.246 &	0.415 \\\hline
    \end{tabular}
   
\end{table*}

\begin{figure*}
\begin{minipage}{.33\textwidth}
    \centering
    \includegraphics[width=1\linewidth]{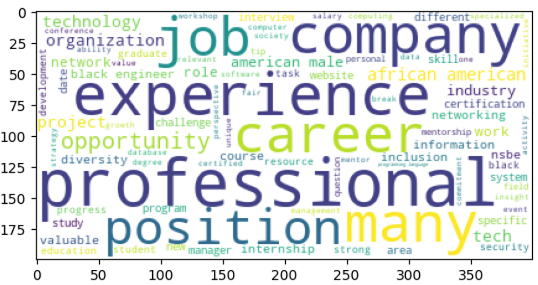}  
    {Profile: M-AA-J-CS}
\end{minipage}
\begin{minipage}{.33\textwidth}
     \centering\includegraphics[width=1\linewidth]{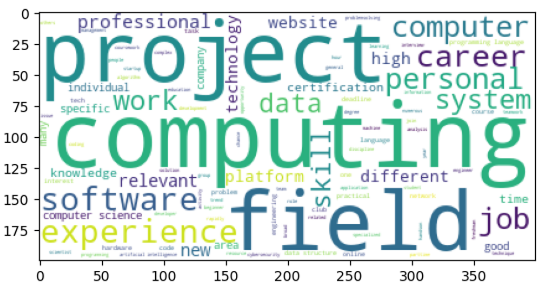}  
   {Profile: M-W-F-U}
\end{minipage}
\begin{minipage}{.33\textwidth}
   \centering
    \includegraphics[width=1\linewidth]{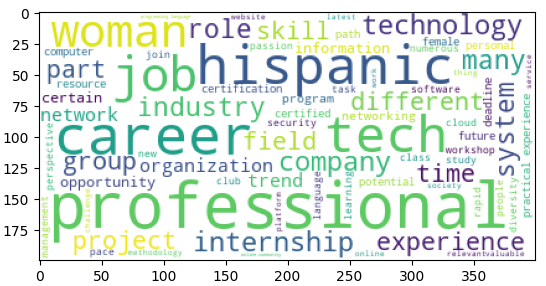} 
    {Profile: F-H-F-CS}
  \end{minipage}
\caption{Word Clouds of All question answers for profile M-AA-J-CS, M-W-F-U, and F-H-F-CS using GPT-4o}
\label{gpt4}
\end{figure*}

\begin{figure*}
\begin{minipage}{.33\textwidth}
    \centering
    \includegraphics[width=1\linewidth]{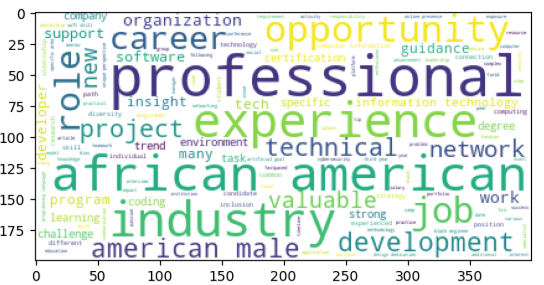}  
    {Profile: M-AA-J-CS}
\end{minipage}
\begin{minipage}{.33\textwidth}
     \centering\includegraphics[width=1\linewidth]{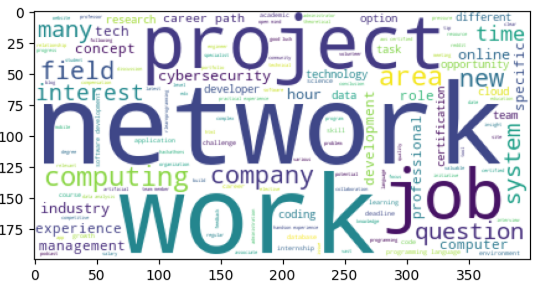}  
   {Profile: M-W-F-U}
\end{minipage}
\begin{minipage}{.33\textwidth}
   \centering
    \includegraphics[width=1\linewidth]{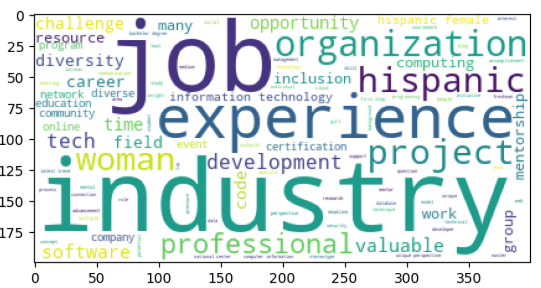} 
    {Profile: F-H-F-CS}
  \end{minipage}
\caption{Word Clouds of All question answers for profile M-AA-J-CS, M-W-F-U, and F-H-F-CS using LLaMA 3}
\label{llama3}
\end{figure*}

\begin{figure*}
\begin{minipage}{.33\textwidth}
    \centering
    \includegraphics[width=1\linewidth]{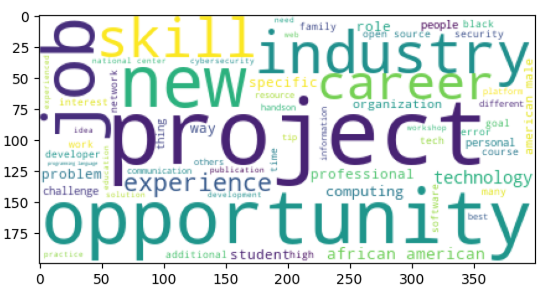}  
    {Profile: M-AA-J-CS}
\end{minipage}
\begin{minipage}{.33\textwidth}
     \centering\includegraphics[width=1\linewidth]{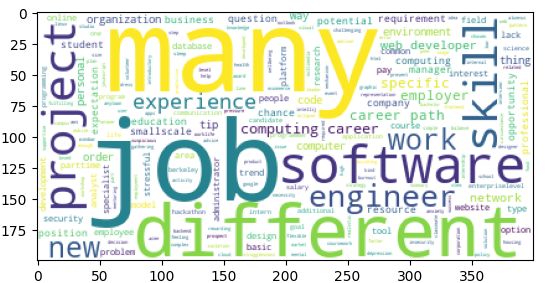}  
   {Profile: M-W-F-U}
\end{minipage}
\begin{minipage}{.33\textwidth}
   \centering
    \includegraphics[width=1\linewidth]{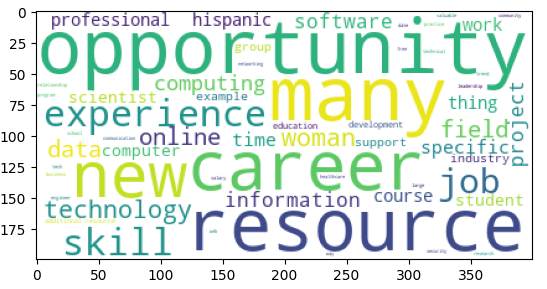} 
    {Profile: F-H-F-CS}
  \end{minipage}
\caption{Word Clouds of All question answers for profile M-AA-J-CS, M-W-F-U, and F-H-F-CS using Palm 2}
\label{palm2}
\end{figure*}

\section{Experimental Settings and Evaluation Results}
\subsection{Result of NLP Analysis}
Table IV show the results of semantic similarity analysis using the method described in Section III.C using three LLMs, respectively. The values in Table III display the percentage of sentences in a response that share the same semantic meaning with sentences from other responses, yet originate from different student profiles. Based on the overall average of the semantic similarity analysis, we found that the answer to the male white freshman student has less overlapped content with other two students. After investigating the answers, we found this is primarily because that this student is an undecided major students. So that the answer to the mentoring questions are less specific, but more in general with regards to how to explore various subjects or opportunities to discover personal interests. Whereas the answer to the male African American junior student and the female, Hispanic freshman student have more overlapped content. Since they both are in the computer science major, the answers often include the same content that provides to computer science majors, such as a set of specific certifications that students can acquire for career planning etc.   

Upon comparing responses to questions that require more personalized content, we found that GPT-4o and Palm 2 tend to produce answers with higher average similarity scores to generic questions. This trend is not observed with LLaMA 3. For instance, for the question, `How many years of education/training should I expect to get a computing job?', LLaMA 3 provides significantly varied answers based on three different profiles, while GPT-4o and Palm 2 offer responses with slightly higher similarity. Further investigation revealed that LLaMA 3 personalizes its answers according to the professional level of the students. If a student has chosen computer science as their major, LLaMA 3 tailors its response based on whether the student is a junior or freshman, indicating the number of additional years needed to secure a job. For students who have not decided on a major, LLaMA 3 provides detailed information on the range of options, from an associate degree to a PhD, including the corresponding number of years required for each.

Comparing the answers from the three LLMs for each student profile, we found that they also include some shared content that is generic to the questions. For example, in response to the question, `What I want to know about a computing career is whether many jobs are available and how much positions in this field pay,' all three LLMs reference labor statistics data to provide detailed information about the average salaries for various computing jobs.

To summarize the unique aspects of the answers to the questions based on each profile, we generated word cloud to highlight the most unique aspects in the answers as accumulative frequency of the words in the answers of all questions. Figure \ref{gpt4} to \ref{palm2} show the word clouds generated for profiles  M-AA-J-CS, M-W-F-U, and F-H-F-CS, respectively. Based on the word clouds we can tell that for profile M-AA-J-CS, all three LLMs mentioned `African American' to some extend, whereas LLaMA 3 emphasized more and also mentioned male. Since this is a junior CS major student, many answers mentioned `professional network', `professional profile', `industry experience' for `job' and `career'. Different from the answers for profile 1, the answers for profile 2 who is an undecided major freshman student, the answers from all LLMs emphasize on working on `project' to know `computing field'  and gain `skills'. None of these answers specifically mentioned about how to handle diversity challenges etc. For profile 3 who is a Hispanic female CS major freshman, both GPT-4o and LLaMA 3 mentioned more about `Hispanic' `woman' in the answers of the questions, which mean these two LLMs try to personalize the answers towards the profile of the mentee, whereas Palm 2 personalized less in the content of the answers. However, the answers for profile 3 shared some keywords with the answers for profile 1, such as `professional', `career', `experience' since they both are CS majors, the answers for them often contain `professional certification', `professional skills', `career advancement', `career services', `career path', `personal experience', etc., in different answers.

\subsection{Result of Human Evaluation}
Figure \ref{fig:mentoringhuman} presents the results of the human evaluation based on a survey using the evaluation questions listed in Table III. GPT-4o received the highest human evaluation scores for most of the survey questions, while Palm 2 scored the lowest. For three questions — `Does the answer recommend competency developments?', `Does the answer promote professional development and networking?', and `Is the answer respectful?' — both evaluators gave the highest scores across all LLMs. This indicates that all three LLMs effectively identified information relevant to competency development in computing domains, promoted professional networking to enhance soft skills, and provided answers in a respectful manner. However, the human evaluation revealed that all three LLMs performed poorly in helping students set appropriate goals. Upon reviewing all the questions and answers, we found that this may be due to the absence of questions specifically addressing personal or career goals. While some answers mentioned general pathways to achieve career goals, they lacked specificity to the students' profiles. Therefore, the lower scores may not accurately reflect the actual capabilities of the LLMs. Our research primarily focuses on the question-answering aspects of AI mentoring. However, further investigation is needed into interactive AI mentoring in a chatbot format.


\section{Conclusion and Future work}

Our research evaluated the ability of LLMs to provide personalized mentoring by considering students' social backgrounds and skill levels. The findings indicate that while LLMs can account for personal backgrounds to some extent, human involvement is essential to deliver truly personalized mentoring experiences. We also suggest that fine-tuning LLMs with educational data could enhance their performance. However, ethical concerns, particularly regarding privacy, bias, and transparency, must be addressed for real-world implementation of such systems.

Future work involves developing a human-in-the-loop mentoring system using LLMs, incorporating a chatbot-like mechanism to gain a deeper understanding of individual needs and provide tailored mentoring experiences. This approach can also help mitigate biases and unfairness that may arise from the LLMs' training data.


\bibliographystyle{plain}
\bibliography{ref}

\begin{thebibliography}{10}

\bibitem{abd2023large}
Alaa Abd-Alrazaq, Rawan AlSaad, Dari Alhuwail, Arfan Ahmed, Padraig~Mark Healy, Syed Latifi, Sarah Aziz, Rafat Damseh, Sadam~Alabed Alrazak, Javaid Sheikh, et~al.
\newblock Large language models in medical education: opportunities, challenges, and future directions.
\newblock {\em JMIR Medical Education}, 9(1):e48291, 2023.

\bibitem{achiam2023gpt}
Josh Achiam, Steven Adler, Sandhini Agarwal, Lama Ahmad, Ilge Akkaya, Florencia~Leoni Aleman, Diogo Almeida, Janko Altenschmidt, Sam Altman, Shyamal Anadkat, et~al.
\newblock Gpt-4 technical report.
\newblock {\em arXiv preprint arXiv:2303.08774}, 2023.

\bibitem{akiba2023ai}
Daisuke Akiba and Michelle~C Fraboni.
\newblock Ai-supported academic advising: Exploring chatgpt’s current state and future potential toward student empowerment.
\newblock {\em Education Sciences}, 13(9):885, 2023.

\bibitem{anil2023palm}
Rohan Anil, Andrew~M Dai, Orhan Firat, Melvin Johnson, Dmitry Lepikhin, Alexandre Passos, Siamak Shakeri, Emanuel Taropa, Paige Bailey, Zhifeng Chen, et~al.
\newblock Palm 2 technical report.
\newblock {\em arXiv preprint arXiv:2305.10403}, 2023.

\bibitem{mithun2024designing}
Anonymous.
\newblock Anonymous.
\newblock In {\em 2024 ASEE Annual Conference \& Exposition}, 2024.

\bibitem{bagai2024designing}
Rahul Bagai and Vaishali Mane.
\newblock Designing an ai-powered mentorship platform for professional development: opportunities and challenges.
\newblock {\em arXiv preprint arXiv:2407.20233}, 2024.

\bibitem{benitez2024harnessing}
Trista~M Ben{\'\i}tez, Yueyuan Xu, J~Donald Boudreau, Alfred Wei~Chieh Kow, Fernando Bello, Le~Van~Phuoc, Xiaofei Wang, Xiaodong Sun, Gilberto Ka-Kit Leung, Yanyan Lan, et~al.
\newblock Harnessing the potential of large language models in medical education: promise and pitfalls.
\newblock {\em Journal of the American Medical Informatics Association}, 31(3):776--783, 2024.

\bibitem{berk2005measuring}
Ronald~A Berk, Janet Berg, Rosemary Mortimer, Benita Walton-Moss, and Theresa~P Yeo.
\newblock Measuring the effectiveness of faculty mentoring relationships.
\newblock {\em Academic medicine}, 80(1):66--71, 2005.

\bibitem{biswas2023role}
Som Biswas.
\newblock Role of chatgpt in computer programming.
\newblock {\em Mesopotamian Journal of Computer Science}, 2023:9--15, 2023.

\bibitem{buruk2023academic}
O{\u{g}}uz'Oz' Buruk.
\newblock Academic writing with gpt-3.5 (chatgpt): reflections on practices, efficacy and transparency.
\newblock In {\em Proceedings of the 26th International Academic Mindtrek Conference}, pages 144--153, 2023.

\bibitem{cronje2023exploring}
Johannes Cronj{\'e}.
\newblock Exploring the role of chatgpt as a peer coach for developing research proposals: Feedback quality, prompts, and student reflection.
\newblock {\em Electronic Journal of e-Learning}, 22(2):1--15, 2023.

\bibitem{dubey2024llama}
Abhimanyu Dubey, Abhinav Jauhri, Abhinav Pandey, Abhishek Kadian, Ahmad Al-Dahle, Aiesha Letman, Akhil Mathur, Alan Schelten, Amy Yang, Angela Fan, et~al.
\newblock The llama 3 herd of models.
\newblock {\em arXiv preprint arXiv:2407.21783}, 2024.

\bibitem{fagbohun2024beyond}
O~Fagbohun, NP~Iduwe, M~Abdullahi, A~Ifaturoti, and OM~Nwanna.
\newblock Beyond traditional assessment: Exploring the impact of large language models on grading practices.
\newblock {\em Journal of Artifical Intelligence and Machine Learning \& Data Science}, 2(1):1--8, 2024.

\bibitem{frankford2024ai}
Eduard Frankford, Clemens Sauerwein, Patrick Bassner, Stephan Krusche, and Ruth Breu.
\newblock Ai-tutoring in software engineering education.
\newblock In {\em Proceedings of the 46th International Conference on Software Engineering: Software Engineering Education and Training}, pages 309--319, 2024.

\bibitem{gavsevic2015let}
Dragan Ga{\v{s}}evi{\'c}, Shane Dawson, and George Siemens.
\newblock Let’s not forget: Learning analytics are about learning.
\newblock {\em TechTrends}, 59:64--71, 2015.

\bibitem{goodman2023accuracy}
Rachel~S Goodman, J~Randall Patrinely, Cosby~A Stone, Eli Zimmerman, Rebecca~R Donald, Sam~S Chang, Sean~T Berkowitz, Avni~P Finn, Eiman Jahangir, Elizabeth~A Scoville, et~al.
\newblock Accuracy and reliability of chatbot responses to physician questions.
\newblock {\em JAMA network open}, 6(10):e2336483--e2336483, 2023.

\bibitem{haindl2024students}
Philipp Haindl and Gerald Weinberger.
\newblock Students’ experiences of using chatgpt in an undergraduate programming course.
\newblock {\em IEEE Access}, 12:43519--43529, 2024.

\bibitem{hartley2024artificial}
Kendall Hartley, Merav Hayak, and Un~Hyeok Ko.
\newblock Artificial intelligence supporting independent student learning: An evaluative case study of chatgpt and learning to code.
\newblock {\em Education Sciences}, 14(2):120, 2024.

\bibitem{holstein2018classroom}
Kenneth Holstein, Gena Hong, Mera Tegene, Bruce~M McLaren, and Vincent Aleven.
\newblock The classroom as a dashboard: Co-designing wearable cognitive augmentation for k-12 teachers.
\newblock In {\em Proceedings of the 8th international conference on learning Analytics and knowledge}, pages 79--88, 2018.

\bibitem{ismail2019lecturer}
Mohammed Ismail and Abejide Ade-Ibijola.
\newblock Lecturer's apprentice: A chatbot for assisting novice programmers.
\newblock In {\em 2019 international multidisciplinary information technology and engineering conference (IMITEC)}, pages 1--8. IEEE, 2019.

\bibitem{kosar2024computer}
Toma{\v{z}} Kosar, Dragana Ostoji{\'c}, Yu~David Liu, and Marjan Mernik.
\newblock Computer science education in chatgpt era: Experiences from an experiment in a programming course for novice programmers.
\newblock {\em Mathematics}, 12(5):629, 2024.

\bibitem{kumar2010architecture}
Rohit Kumar and Carolyn~P Rose.
\newblock Architecture for building conversational agents that support collaborative learning.
\newblock {\em IEEE Transactions on Learning Technologies}, 4(1):21--34, 2010.

\bibitem{lehman2021growing}
Kathleen~J Lehman, Julia~Rose Karpicz, Veronika Rozhenkova, Jamelia Harris, and Tomoko~M Nakajima.
\newblock Growing enrollments require us to do more: Perspectives on broadening participation during an undergraduate computing enrollment boom.
\newblock In {\em Proceedings of the 52nd ACM Technical Symposium on Computer Science Education}, pages 809--815, 2021.

\bibitem{limo2023personalized}
Fernando Antonio~Flores Limo, David Raul~Hurtado Tiza, Maribel~Mamani Roque, Edward~Espinoza Herrera, Jos{\'e} Patricio~Mu{\~n}oz Murillo, Jorge~Jinchu{\~n}a Huallpa, Victor Andre~Ariza Flores, Alejandro Guadalupe~Rinc{\'o}n Castillo, Percy Fritz~Puga Pe{\~n}a, Christian Paolo~Martel Carranza, et~al.
\newblock Personalized tutoring: Chatgpt as a virtual tutor for personalized learning experiences.
\newblock {\em Przestrze{\'n} Spo{\l}eczna (Social Space)}, 23(1):293--312, 2023.

\bibitem{okonkwo2022revision}
Chinedu~Wilfred Okonkwo and Abejide Ade-Ibijola.
\newblock Revision-bot: A chatbot for studying past questions in introductory programming.
\newblock {\em IAENG International Journal of Computer Science}, 49(3), 2022.

\bibitem{pankiewicz2023large}
Maciej Pankiewicz and Ryan~S Baker.
\newblock Large language models (gpt) for automating feedback on programming assignments.
\newblock {\em arXiv e-prints}, pages arXiv--2307, 2023.

\bibitem{rane2024enhancing}
Nitin Rane.
\newblock Enhancing the quality of teaching and learning through gemini, chatgpt, and similar generative artificial intelligence: Challenges, future prospects, and ethical considerations in education.
\newblock {\em TESOL and Technology Studies}, 5(1):1--6, 2024.

\bibitem{roll2016evolution}
Ido Roll and Ruth Wylie.
\newblock Evolution and revolution in artificial intelligence in education.
\newblock {\em International journal of artificial intelligence in education}, 26:582--599, 2016.

\bibitem{sanya2024chatgpt}
Frank Sanya-Isijola and Fok-Han Leung.
\newblock Chatgpt: bridging the gap on mentorship for international medical graduates in low enrolment specialties.
\newblock {\em Canadian Medical Education Journal}, 15(2):93, 2024.

\bibitem{supriyadi2023exploring}
Edi Supriyadi and KS~Kuncoro.
\newblock Exploring the future of mathematics teaching: Insight with chatgpt.
\newblock {\em Union: Jurnal Ilmiah Pendidikan Matematika}, 11(2):305--316, 2023.

\bibitem{thomasusing}
Danielle~R Thomas, Erin Gatz, Shivang Gupta, Jionghao Lin, Cindy Tipper, and Kenneth Koedinger.
\newblock Using generative ai to provide feedback to adult tutors in training and assess real-life performance.

\bibitem{wang2023exploring}
Tianjia Wang, Daniel~Vargas D{\'\i}az, Chris Brown, and Yan Chen.
\newblock Exploring the role of ai assistants in computer science education: Methods, implications, and instructor perspectives.
\newblock In {\em 2023 IEEE Symposium on Visual Languages and Human-Centric Computing (VL/HCC)}, pages 92--102. IEEE, 2023.

\bibitem{wardat2023chatgpt}
Yousef Wardat, Mohammad~A Tashtoush, Rommel AlAli, and Adeeb~M Jarrah.
\newblock Chatgpt: A revolutionary tool for teaching and learning mathematics.
\newblock {\em Eurasia Journal of Mathematics, Science and Technology Education}, 19(7):em2286, 2023.

\bibitem{xu2024large}
Hanyi Xu, Wensheng Gan, Zhenlian Qi, Jiayang Wu, and Philip~S Yu.
\newblock Large language models for education: A survey.
\newblock {\em arXiv preprint arXiv:2405.13001}, 2024.

\bibitem{xue2024does}
Yuankai Xue, Hanlin Chen, Gina~R Bai, Robert Tairas, and Yu~Huang.
\newblock Does chatgpt help with introductory programming? an experiment of students using chatgpt in cs1.
\newblock In {\em Proceedings of the 46th International Conference on Software Engineering: Software Engineering Education and Training}, pages 331--341, 2024.

\bibitem{yilmaz2023augmented}
Ramazan Yilmaz and Fatma Gizem~Karaoglan Yilmaz.
\newblock Augmented intelligence in programming learning: Examining student views on the use of chatgpt for programming learning.
\newblock {\em Computers in Human Behavior: Artificial Humans}, 1(2):100005, 2023.

\bibitem{yilmaz2023effect}
Ramazan Yilmaz and Fatma Gizem~Karaoglan Yilmaz.
\newblock The effect of generative artificial intelligence (ai)-based tool use on students' computational thinking skills, programming self-efficacy and motivation.
\newblock {\em Computers and Education: Artificial Intelligence}, 4:100147, 2023.

\bibitem{yukawa2020new}
Michi Yukawa, Stuart~A Gansky, Patricia O’Sullivan, Arianne Teherani, and Mitchell~D Feldman.
\newblock A new mentor evaluation tool: Evidence of validity.
\newblock {\em PloS one}, 15(6):e0234345, 2020.

\bibitem{zohery2023chatgpt}
Medhat Zohery.
\newblock Chatgpt in academic writing and publishing: A comprehensive guide.
\newblock {\em Artificial intelligence in academia, research and science: ChatGPT as a case study}, pages 10--61, 2023.

\end{thebibliography}

\end{document}